\documentclass[conference]{IEEEtran}
\IEEEoverridecommandlockouts

\usepackage{cite}
\usepackage{amsmath,amssymb,amsfonts}
\usepackage{algorithmic}
\usepackage{graphicx}
\usepackage{textcomp}
\usepackage{xcolor}
\usepackage{url}
\usepackage{enumitem}
\usepackage{hyperref}
\usepackage{caption}
\usepackage{subcaption}
\usepackage{tikzpagenodes}

\def\BibTeX{{\rm B\kern-.05em{\sc i\kern-.025em b}\kern-.08em
    T\kern-.1667em\lower.7ex\hbox{E}\kern-.125emX}}
    
\makeatletter



\begin{document}
\title{Interpretable Multi Labeled Bengali Toxic Comments Classification using Deep Learning}

\author{\IEEEauthorblockN{Tanveer Ahmed Belal}
\IEEEauthorblockA{\textit{Islamic University of Technology}\\
Gazipur, Bangladesh  \\
tanveer@iut-dhaka.edu}
\and
\IEEEauthorblockN{G. M. Shahariar}
\IEEEauthorblockA{\textit{Bangladesh University of Engineering}\\
\textit{and Technology}\\
Dhaka, Bangladesh \\
sshibli745@gmail.com}
\and
\IEEEauthorblockN{Md. Hasanul Kabir}
\IEEEauthorblockA{\textit{Islamic University of Technology}\\
Gazipur, Bangladesh \\
hasanul@iut-dhaka.edu}}

\maketitle

\scalebox{0.85}{
\begin{tikzpicture}[remember picture,overlay]
    \node[align=center,text=blue] at ([xshift=4.7em, yshift=-14em]current page text area.south) {Accepted and presented at ``2023 3rd International Conference on Electrical, Computer and Communication Engineering (ECCE)''.\\
© 2023 IEEE. Personal use of this material is permitted. Permission from IEEE must be obtained for all other uses, in any current \\or future media, including reprinting/republishing this material for advertising or promotional purposes, creating new collective works,\\ for resale or redistribution to servers or lists, or reuse of any copyrighted component of this work in other works.};
\end{tikzpicture}}

\begin{abstract}
This paper presents a deep learning-based pipeline for categorizing Bengali toxic comments, in which at first a binary classification model is used to determine whether a comment is toxic or not, and then a multi-label classifier is employed to determine which toxicity type the comment belongs to. For this purpose, we have prepared a manually labeled dataset consisting of $16,073$ instances among which $8,488$ are \textit{Toxic} and any toxic comment may correspond to one or more of the six toxic categories – \textit{vulgar, hate, religious, threat, troll,} and \textit{insult} simultaneously.  \textit{Long Short Term Memory (LSTM)} with \textit{BERT Embedding} achieved $89.42\%$ accuracy for the binary classification task while as a multi-label classifier, a combination of \textit{Convolutional Neural Network} and \textit{Bi-directional Long Short Term Memory (CNN-BiLSTM)} with \textit{attention} mechanism achieved   $78.92\%$ accuracy and $0.86$ as weighted F1-score. To explain the predictions and interpret the word feature importance during classification by the proposed models, we utilized \textit{Local Interpretable Model-Agnostic Explanations (LIME)} framework. We have made our dataset public and can be accessed at – \url{https://github.com/deepu099cse/Multi-Labeled-Bengali-Toxic-Comments-Classification}
\end{abstract}


\section{Introduction}
The growth of mass media and social networks allows people to share their thoughts and emotions more easily than ever before. This allows information to spread more rapidly than before. However, the rise in social media usage has also led to worries about the spread of harmful content under the guise of the ability to communicate freely. Toxic content is anything virtual that makes a person feel uncomfortable or hurt. This term is somewhat ambiguous since what one person deems to be problematic may not be viewed that way by another. Sexually explicit and violent; objectionable information; hate speech; political or religious content; cyberbullying; and harassment are some of the characteristics of toxic content. Toxic content is more audible on digital platforms since it catches the attention of the public. 

Though Bengali is the world's seventh most frequently spoken language, with over $229$ million native speakers \cite{b1}, its rapid usage as a profane and abusive language in case of  harassment, cyberbullying, and defamation crime poses a barrier to access and entertainment, even to the point of engaging on social networks. Women in Bangladesh who reported suffering online harassment in $2019$ made up seventy percent of the individuals between the ages of fifteen and twenty five. Eighteen percent of harassment complaints and cases involving cyberbullying and defamation were processed by the nation's sole cybercrime tribunal \cite{b2}. Numerous suicides have already been caused by the toxic social media comments that were used in cyberbullying situations throughout the years.

Several research works have been conducted to deal with toxic content in Bengali language such as \textit{toxic comment detection} \cite{b3}, \textit{abusive comment detection} \cite{b4}, \textit{cyberbullying detection} \cite{b5}, \textit{hate speech detection} \cite{b6}. The problem with majority of these works is that the amount of data they used is inadequate and in most cases the datasets were unbalanced. Moreover, most of the research works considered toxic comment classification as a binary or multi-class classification problem. But it is quite possible that a toxic comment can fall into multiple classes simultaneously. Furthermore, we observed that word embedding techniques like word2vec, fasttext, GloVe were frequently used for feature representation which is problematic with respect to context of the sentence, missing words etc. Although these works leveraged deep learning and transformer based models for classification tasks but the models were not interpretable. To address these issues, this paper proposes a two stage pipeline that leverages deep learning and transformer based models to identify toxic comments in Bengali by formulating toxicity detection as a multi-label classification problem. In summary, the following are the primary contributions we made in this paper:  
 
\begin{itemize}
\item We have prepared a Bengali toxic comment detection dataset with $16,073$ instances in total. Among them $7,585$ are \textit{Non-toxic} and $8,488$ are \textit{Toxic}. Toxic instances are manually labeled in one or more of the six classes – \textit{vulgar, hate, religious, threat, troll,} and \textit{insult}. 
\item We have proposed a pipeline where we first identify whether a comment is toxic or non-toxic by using a binary classification model, and if identified as toxic, we then employ a multi-label classifier to categorize the toxicity type.
\item We report that \textit{LSTM with BERT Embedding} achieved highest $89.42\%$ accuracy as binary classifier while for multi-label classification, \textit{CNN-BiLSTM with attention} performed best because the attention layer selectively focused on the important portions of the input sequence which helped the CNN-BiLSTM model to better understand the correlation among them.  
\item Finally, we have utilized a text explainer framework (LIME) to interpret the predictions of the deep learning models .
\end{itemize}

\begin{figure*}[ht]
\centerline{\includegraphics[scale=0.8]{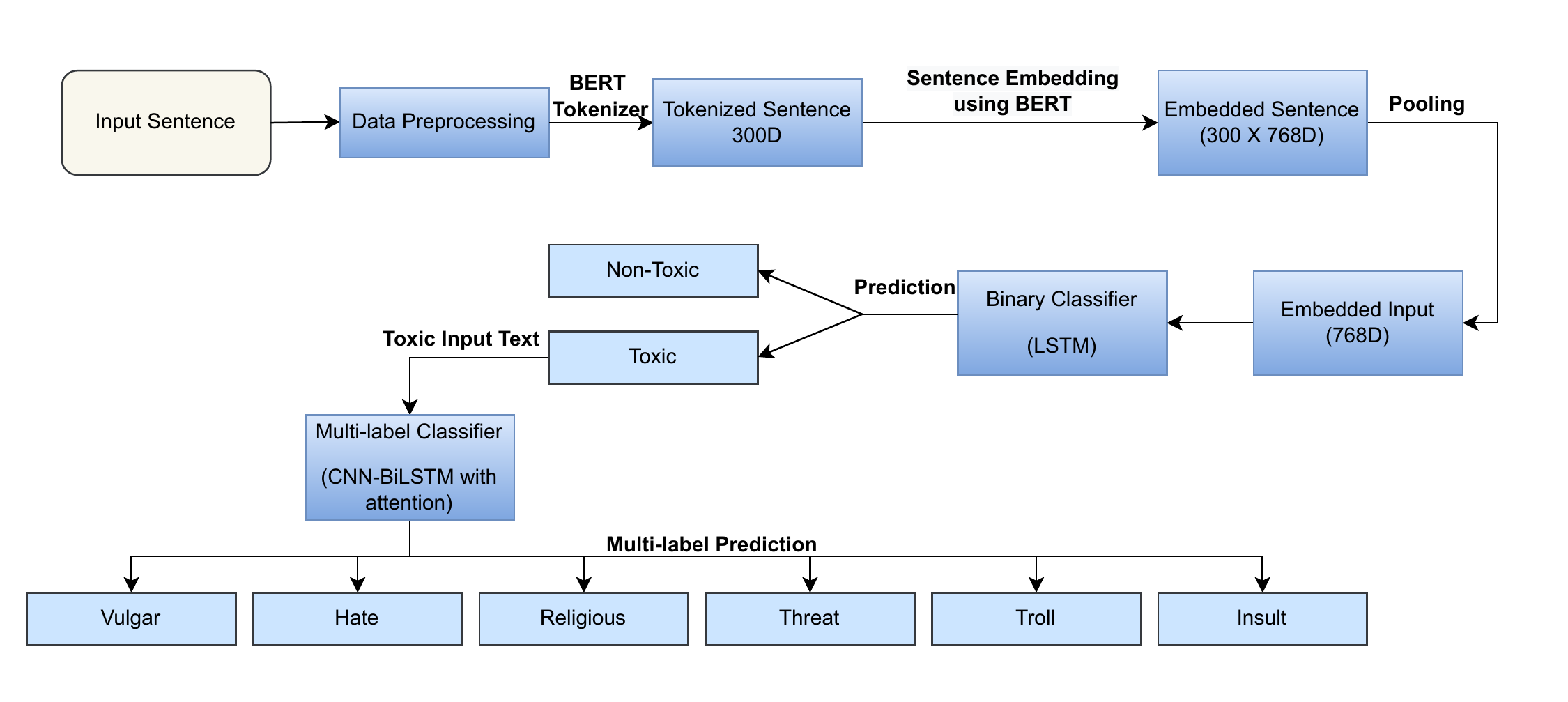}}
\caption{Proposed methodology for multi-labeled toxic comment classification.}
\label{fig:methodology}
\vspace{-4mm}
\end{figure*}

\section{Related Works}
Based on the problem formulation, the previous works on toxicity detection in Bengali language can be divided into two broader categories: Binary and multi-class classification.

\noindent\textbf{Binary Classification}: Banik et al. \cite{b3} used both supervised machine learning models such as \textit{Naïve Bayes}, \textit{Support Vector Machine}, \textit{Logistic Regression} and deep learning models such as \textit{Convolutional Neural Network}, \textit{Long Short Term Memory} to classify Bengali comments as toxic and non-toxic. \textit{Convolutional Neural Network} achieved the highest accuracy of $95.30\%$ for binary classification. For word embedding they used \textit{word2vec}. Chakraborty et al. \cite{b7} classified threat and abusive Bengali texts using both traditional machine learning classifiers such as \textit{Multinomial Naive Bayes}, \textit{Support Vector Machine} and deep learning classifiers such as \textit{Convolutional Neural Network} with \textit{Long Short Term Memory (CNN-LSTM)} considering the problem as a binary classification task. They reported the maximum $78\%$ accuracy using SVM with linear kernel. Ahammed et al. \cite{b8} utilized two machine learning models- \textit{Support Vector Machine} and \textit{Naïve Bayes} to detect hate speech and reported $72\%$ accuracy using SVM where Bengali data were collected form Facebook post. To detect toxic speech, Malik et al. \cite{b9} used a bunch of machine learning models such as \textit{Logistic Regression, Support Vector Machine, Decision Tree, Random Forest, XGBoost} and deep learning algorithms: \textit{Convolutional Neural Network, Multi Layer Perceptron, Long Short Term Memory}. They reported $82\%$ accuracy with CNN and for word embedding they used \textit{BERT} and \textit{fasttext}. 

\noindent\textbf{Multiclass Classification}: Faisal et al. \cite{b5} used \textit{Convolutional Neural Network} with \textit{Long Short Term Memory (CNN-LSTM)} to detect cyberbullying and reported $87.91\%$ accuracy for binary classification and for multi-class classification they reported $85\%$ accuracy using ensemble technique with the help of binary classifier. Das et al. \cite{b10} used \textit{Convolutional Neural Network} with \textit{Long Short Term Memory, Convolutional Neural Network with Gated Recurrent Unit} and \textit{attention based Convolutional Neural Network} to detect Bengali hate speech and \textit{attention-based Convolutional Neural Network} approach achieved $77\%$ accuracy. Karim et al. \cite{b6} detected Bengali hate speech using \textit{Multichannel Convolutional-Long Short Term Memory (MConv-LSTM)} and for embedding they used \textit{BangFastText, Word2Vec, GloVe}. Their model achieved $92.30\%$ F1-score with BangFastText embedding. Aurpa et al. \cite{b4} used pre-training language architectures, \textit{BERT} and \textit{ELECTRA} for identifying abusive Bengali comments. BERT model performed notably better than ELECTRA which achieved a maximum accuracy $85\%$.

\section{Dataset Preparation}
\subsection{Data Accumulation}
The text samples were gathered from three sources: the multi-labeled Bangla-Abusive-Comment-Dataset \cite{b14} and the multi-class Bengali Hate Speech Dataset \cite{b6} and Bangla Online Comments Dataset \cite{b5}. Upon careful examination, it was determined that the original labeling of the texts in these datasets was not accurate or consistent. In some cases, the texts were deemed to belong in multiple categories at once, leading to the decision to manually categorize them into six classes: \textit{vulgar, hate, religious, threat, troll, insult}, where each text could be assigned multiple labels. Reclassifying the texts from multiple datasets into a new set of categories is crucial for several reasons, including ensuring consistency, enhancing data quality, providing more insightful information about the nature and extent of toxicity, boosting the performance of machine learning models, and streamlining data management. The reclassification also offers a clearer understanding of the toxicity present in the data and improves the accuracy of machine learning models.

\begin{table}[h]
\centering
\caption{Statistics of the toxic comments for multi-label classification}
\label{tab:dataset}
\begin{tabular}{|c|c|c|} 
\hline
\textbf{Class} & \begin{tabular}[c]{@{}c@{}}\textbf{No. of }\\\textbf{ instances}\end{tabular} & \begin{tabular}[c]{@{}c@{}}\textbf{Kappa}\\\textbf{Score}\end{tabular}  \\ 
\hline
vulgar         & 2505                                                                          & 0.88                                                                    \\ 
\hline
hate           & 1898                                                                          & 0.79                                                                    \\ 
\hline
religious      & 1418                                                                          & 0.84                                                                    \\ 
\hline
threat         & 1419                                                                          & 0.71                                                                    \\ 
\hline
troll          & 1643                                                                          & 0.66                                                                    \\ 
\hline
insult         & 2719                                                                          & 0.74                                                                    \\
\hline
\end{tabular}
\vspace{-7mm}
\end{table}

\subsection{Data Annotation}
For labeling the data, we followed the procedure mentioned in \cite{b11} and sought the assistance of three professional annotators. Each instance in the dataset was labeled by two annotators, and if any disagreement arose, a third expert was consulted to resolve the matter through discussion and debate. To validate and assess the quality of the annotations, we evaluated the inter-annotator agreement using the Cohen's kappa coefficient \cite{b12}. We evaluated the annotators' trustworthiness by selecting $100$ random sentences (previously labeled by an expert) and creating $30$ control samples. All three annotators had a trustworthiness score above $80\%$ based on the evaluation. The control samples were easy to understand and unknown to the annotators. The average kappa score for both \textit{Toxic} and \textit{Non-toxic} classes was $0.96$, and the kappa scores for each individual toxic category can be seen in Table \ref{tab:dataset}.

\subsection{Dataset statistics}
The dataset consists of a total of $16,073$ instances, with $7,585$ instances being labeled as \textit{Non-toxic} and $8,488$ instances labeled as \textit{Toxic}. The distribution of toxic instances per class is displayed in Table \ref{tab:dataset}.

\section{Proposed Methodology}
The proposed system for toxic comment classification is depicted in figure \ref{fig:methodology}. At first, we consider the problem as a Binary classification task where we detect whether a input text is \textit{Toxic} or \textit{Non-toxic}. If the input text is identified as \textit{Toxic} then we consider a multi-label classification problem to categorize the text into one or more of the six classes – \textit{vulgar, hate, religious, threat, troll,} and \textit{insult}. The key phases of the proposed method are further detailed below.
\begin{enumerate}[label=\textbf{Step \arabic{enumi})}, wide, labelwidth=!, labelindent=0pt]
\item \textbf{Input Sentence}: Each raw sentence from the dataset is presented to the proposed model one by one for further processing. 
\item \textbf{Data Preprocessing}: Punctuations, emoticons, hyper-links, and stop words are removed from the input sentence.
\item \textbf{Tokenization}: Each processed sentence is encoded using the \textit{Bangla BERT Tokenizer} which provides input ids and mask ids. Each tokenized sentence gets a length of $300$ tokens and zero padded when required. In the event of length more than $300$, we cut off after $300$. 
\item \textbf{Embedding}: The output of the tokenization step is directly fed to the \textit{Bangla BERT} \cite{b13} which turns each token into a numeric value representation. Each token is embedded by $768$ real values. The output of this step is an embedded vector of size $300*768$.
\item \textbf{Pooling}: To reduce the dimension of the embedding vector $(300*768)$, max pooling is used which provides a real valued vector representation of size $768$ per sentence.
\item\textbf{Binary Classification}: For binary classification, a Long Short Term Memory (LSTM) \cite{b14} network is applied on the embedding vector to learn the dependency of the full sentence. The LSTM layer is followed by max pooling, dropout and dense layer respectively.  In our model, we use leaky rectified linear unit(Leaky ReLU) as the activation function for the hidden layers and finally a sigmoid activation in the dense layer to predict the probability of the input text being in \textit{Toxic} or \textit{Non-toxic} category.
\item\textbf{Multi-label Classification}: If the input text is identified as \textit{Toxic}, then a combination of CNN \cite{b14} and Bidirectional LSTM \cite{b15} model with attention mechanism is utilized to categorize the text into one or more of the six classes – \textit{vulgar, hate, religious, threat, troll,} and \textit{insult}.  The embedding vector of the toxic text is passed into three one dimensional convolutional layers. Each convolutional layer has rectified linear unit (ReLU) as activation function. These three layers consist of $512$, $256$, and $128$ filters with kernel size $4$, $3$ and $2$ respectively. After each convolution layer, we use max pooling with pool size $2$. A Bidirectional LSTM layer (Bi-LSTM) with $128$ units is applied on the output of the last pooling layer. We use L2 regularization in order to overcome the over-fitting problem. We calculate the attention score $\alpha^{A}_{t}$ over the hidden states and concatenate it with the respective hidden state to generate context vector $z^{A}_{t}$ for a given time step $t$. The last dense layer with sigmoid activation function is used to predict the class probabilities.

\begin{equation}
    \alpha^{A}_{t} = softmax(W^{A} H^{A}_{t} + b^{A})
\end{equation}
\begin{equation}
    z^{A}_{t} = \sum_{t = 1} \alpha^{A}_{t} H^{A}_{t}
\end{equation}

\noindent For a given time step $t$, $\alpha^{A}_{t}$ represents the attention score, $W^{A}$ is the filter applied, $b^{A}$ is a bias term, $H^{A}_{t}$ is the concatenation of hidden states both in forward and backward directions, and $z^{A}_{t}$ is the context vector.
\end{enumerate}

\noindent \textbf{Convolutional Neural Network}: Convolution and pooling layers constitute a Convolutional Neural Network (CNN) architecture \cite{b14}. CNN uses artificial neurons to process inputs and convey the outputs. Embedded text is utilized as input. There may be a significant number of hidden layers in a CNN which uses numerous calculations to extract features. Convolution layer serves as the initial step in the feature extraction process from an input. The fully connected layer promotes classification in the output layer. Because information only flows in one direction from inputs to outputs, CNN's are considered as feed forward networks.

\begin{table}[ht]
\centering
\caption{Performance comparison of different Binary Classifiers}
\label{tab:bin-perf}
\begin{tabular}{|c|c|c|c|c|}
\hline
\textbf{Model}                                                         & \textbf{Accuracy} & \textbf{\begin{tabular}[c]{@{}c@{}}Weighted\\ Precision\end{tabular}} & \textbf{\begin{tabular}[c]{@{}c@{}}Weighted\\ Recall\end{tabular}} & \textbf{\begin{tabular}[c]{@{}c@{}}Weighted\\ F1-score\end{tabular}} \\ \hline
\begin{tabular}[c]{@{}c@{}}LSTM + \\ BERT embedding\end{tabular}       & \textbf{89.42\%}              & \textbf{0.89}                                                                  & \textbf{0.89}                                                               & \textbf{0.89}                                                                 \\ \hline
\begin{tabular}[c]{@{}c@{}}MConv-LSTM + \\ BERT embedding\end{tabular} & 87.87\%              & 0.88                                                                  & 0.88                                                               & 0.88                                                                 \\ \hline
\begin{tabular}[c]{@{}c@{}}Bangla BERT \\ fine-tune\end{tabular}       & 88.57\%              & 0.89                                                                  & 0.88                                                               & 0.89                                                                 \\ \hline
\end{tabular}
\vspace{-2mm}
\end{table}

\noindent \textbf{Bidirectional Long Short Term Memory}: Bidirectional Long Short Term Memory (Bi-LSTM) \cite{b15} is one kind of recurrent neural network that can learn long-term data relationships. In contrast to conventional LSTM, the input flows in both ways and uses information from both sides. The cell state in LSTM facilitates the unaltered flow of information through the units by permitting just a few linear interactions. Each unit is equipped with an input, an output, and a forget gate that may add or delete information from the cell state. Using a sigmoid function, the forget gate determines whether information from the previous cell state should be forgotten. In Bi-LSTM an extra LSTM layer is present which reverses the direction of information flow. In short, it indicates that the input sequence flows in reverse in the extra LSTM layer. Finally, the output gate determines which data should be sent to the subsequent concealed state.

\begin{table}[h]
\centering
\caption{Performance comparison of different Multi-label Classifiers}
\label{tab:multi-perf}
\begin{tabular}{|c|c|c|c|c|}
\hline
\textbf{Model}                                                         & \textbf{Accuracy} & \textbf{\begin{tabular}[c]{@{}c@{}}Weighted\\ Precision\end{tabular}} & \textbf{\begin{tabular}[c]{@{}c@{}}Weighted\\ Recall\end{tabular}} & \textbf{\begin{tabular}[c]{@{}c@{}}Weighted\\ F1-score\end{tabular}} \\ \hline
\begin{tabular}[c]{@{}c@{}}LSTM + \\ BERT embedding\end{tabular}       & 76.52\%           & 0.83                                                                  & 0.82                                                               & 0.82                                                                 \\ \hline
\begin{tabular}[c]{@{}c@{}}MConv-LSTM + \\ BERT embedding\end{tabular} & 74.72\%           & 0.83                                                                  & 0.78                                                               & 0.80                                                                 \\ \hline
\begin{tabular}[c]{@{}c@{}}Bangla BERT \\ fine-tune\end{tabular}       & 77.66\%           & 0.86                                                                  & 0.85                                                               & 0.85                                                                \\ \hline
\begin{tabular}[c]{@{}c@{}}CNN-BiLSTM \\ with Attention\end{tabular}   & \textbf{78.92\%}  & \textbf{0.87}                                                                  & \textbf{0.85}                                                               & \textbf{0.86}                                                                 \\ \hline
\end{tabular}
\vspace{-2mm}
\end{table}

\noindent \textbf{Explainable Artificial Intelligence}: Explainable artificial intelligence (XAI) is designed to explain its objective, reasoning, and decision-making process in a way that is understandable to the ordinary person. This feature is crucial to the model's fairness, accountability, and transparency. Local Interpretable Model-agnostic Explanations (LIME) \cite{b17} is a popular text explanation framework. LIME offers locally accurate explanations. It generates $n$ number of samples of the feature vector based on a normal distribution. Using the model prediction whose decisions it is attempting to explain, it derives the target variable for these samples. After getting the data, each row is weighted based on its similarity to the original observation. Then, it employs a feature selection method such as Lasso to retrieve the most essential attributes.

\section{Experimental Results}

\subsection{Experiments}
To compare the performance of our proposed model, we performed seven experiments in total in this work. For binary classification, we performed three experiments: \textit{LSTM with BERT embedding, MConv-LSTM (multichannel CNN with LSTM) with BERT embedding,} and \textit{Bangla BERT fine-tuning}. For multi-label classification, we performed four experiments: \textit{LSTM with BERT embedding, MConv-LSTM with BERT embedding, CNN-BiLSTM with attention mechanism,} and \textit{Bangla BERT fine-tuning}. 


\subsection{Hyper-parameter settings}
For all the experiments, we used $60\%$ of total data as training set and divided the rest data into validation and test set using $60:40$ ratio. All experiments were evaluated on the test set. To split the dataset we used multi-label stratified shuffle split. To train the binary classification models, we considered the six types of toxic classes as \textit{Toxic} class and others as \textit{Non-toxic} class and for training the multi-label classification models, we considered only the six types of toxic data. For all the seven experiments, we used a maximum sequence length of $300$, batch size of $16$, binary cross entropy as loss function, learning rate of $1e-5$, Adam optimizer, and L2 regularization function. The parameter settings of the BERT based model was unchanged. For explaining text instances, we use lime text explainer with $1000$ samples, six and ten word features for binary and multi-label classification respectively. All the experiments were performed on a NVIDIA Tesla K80 GPU.

\begin{table}[h]
\centering
\caption{Per class performance metrics for CNN-BiLSTM with Attention model}
\label{tab:per-class-perf}
\begin{tabular}{|c|c|c|c|c|}
\hline
\textbf{Class} & \textbf{Accuracy} & \textbf{Precision} & \textbf{Recall} & \textbf{F1-score} \\ \hline
Vulgar         & 94.23\%           & 0.87               & 0.94            & 0.91              \\ \hline
Hate           & 91.87\%           & 0.81               & 0.82            & 0.82              \\ \hline
Religious      & 97.48\%           & 0.92               & 0.93            & 0.93              \\ \hline
Threat         & 92.90\%           & 0.78               & 0.79            & 0.79              \\ \hline
Troll          & 94.23\%           & 0.92               & 0.75            & 0.83              \\ \hline
Insult         & 92.61\%           & 0.87               & 0.89            & 0.88              \\ \hline
\end{tabular}
\vspace{-4mm}
\end{table}

\subsection{Evaluation Metrics}
To measure and compare the performance of the experimental models, we utilized average accuracy, weighted precision, weighted recall, weighted f1-score for both binary and multi-label classification. We also measured per class accuracy, precision, recall and f1-score for the proposed multi-label classification model.

\begin{figure*}[h]
		\centering
		\begin{subfigure}[b]{.5\columnwidth}
		    \centering
			\includegraphics[width=1\linewidth]{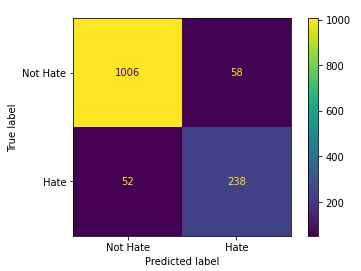}
                \subcaption{hate}
                \label{fig:hate__conmat}
		\end{subfigure}
        \begin{subfigure}[b]{.5\columnwidth}
		    \centering
			\includegraphics[width=1\linewidth]{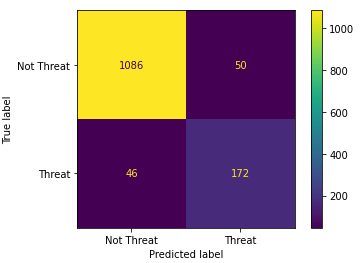}
                 \subcaption{threat}
                \label{fig:threat_conmat}
		\end{subfigure}
		\begin{subfigure}[b]{.5\columnwidth}
		    \centering
			\includegraphics[width=1\linewidth]{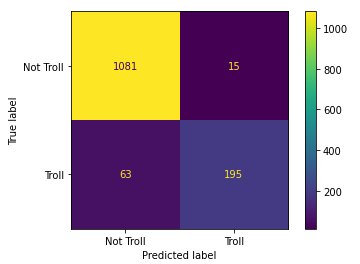}
                \subcaption{troll}
                \label{fig:Comp__conmat}
		\end{subfigure}
        \begin{subfigure}[b]{.5\columnwidth}
		    \centering
			\includegraphics[width=1\linewidth]{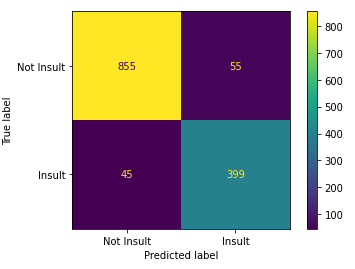}
			\subcaption{insult}
                \label{fig:Bug__conmat}
		\end{subfigure}
		\caption{Confusion matrix for hate, threat, troll and insult category based on per class low f1-scores.}
		\label{fig:visualization_graph}
\vspace{-6mm}
\end{figure*}

\subsection{Experimental Results}
This section deals with the experimental results of all the experiments conducted in this study. For binary classification, three experiments were conducted and the results are presented in table \ref{tab:bin-perf}. From the table, we can observe that LSTM with BERT Embedding achieved highest $89.42\%$ accuracy in categorizing Bengali comments as \textit{Toxic} or \textit{Non-toxic}. For multi-label classification, in total four experiments were conducted in this study. 
\begin{figure}[h]
\centerline{\includegraphics[scale=0.8]{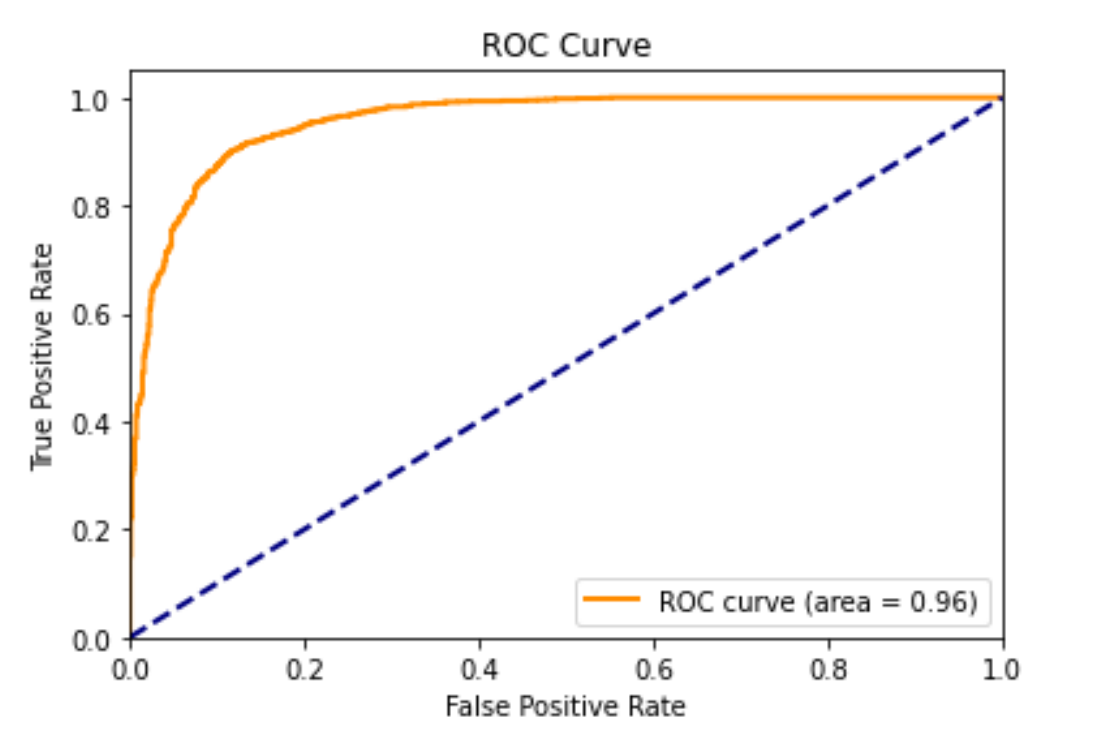}}
\caption{ROC Curve for Binary Classification.}
\label{fig:roc-bin}
\vspace{-2mm}
\end{figure}
Table \ref{tab:multi-perf} represents the corresponding experimental results and CNN-BiLSTM with Attention mechanism achieved the highest overall average accuracy of $78.92\%$ accuracy in categorizing a toxic comment into one or more of the six predefined categories: \textit{vulgar, hate, religious, threat, troll}, and \textit{insult}. As multi-label classification can be considered as a multiple binary classification problem with respect to each class label, to measure and understand the performance of CNN-BiLSTM with Attention model, we report per class accuracy, precision, recall and f1-score in table \ref{tab:per-class-perf}.
\begin{figure}[h]
\centerline{\includegraphics[scale=0.8]{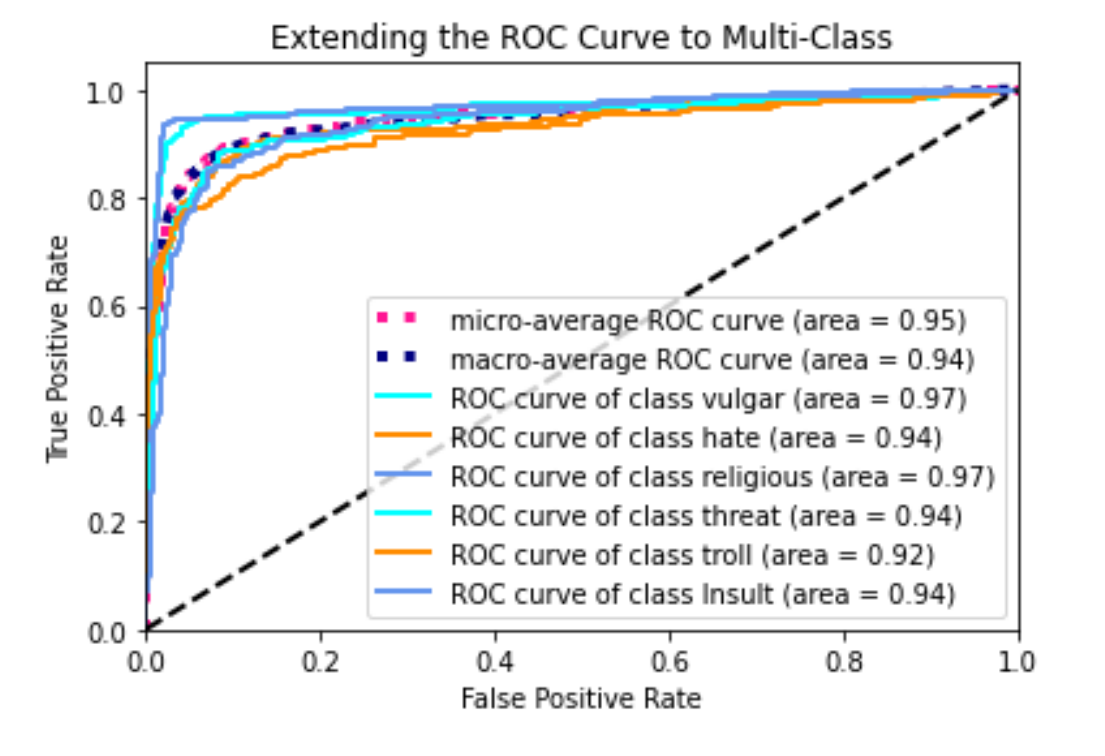}}
\caption{ROC Curve for Multi-label Classification.}
\label{fig:roc-multi}
\vspace{-5mm}
\end{figure}
\subsection{Result Analysis}
LSTM with BERT Embedding performed best with respect to accuracy beating MConv-LSTM with BERT Embedding by approximately $2\%$ and Bangla BERT fine tuning by around $1\%$ in case of binary classification. From table \ref{tab:bin-perf}, we can observe that the difference in performance based on precision, recall, f1-score between MConv-LSTM with BERT Embedding and Bangla BERT fine tuning is minimal as their value ranges from $0.88$ to $0.89$. The performance of the LSTM with BERT Embedding model at all classification thresholds can be visualized in figure \ref{fig:roc-bin}. 
In case of multi label classification, CNN-BiLSTM with Attention mechanism performed best with respect to accuracy beating LSTM with BERT Embedding, MConv-LSTM with BERT Embedding and Bangla BERT fine tuning by approximately $2\%$, $4\%$ and $1\%$ respectively. From table \ref{tab:multi-perf}, we can observe that the difference in performance based on precision, recall, f1-score between Bangla BERT fine tuning and CNN-BiLSTM with Attention mechanism is minimal whereas LSTM with BERT Embedding along with MConv-LSTM with BERT Embedding has difference in performance ranging around $3\%$ to $7\%$.
\begin{figure*}[ht]
\centerline{\includegraphics[width=\textwidth]{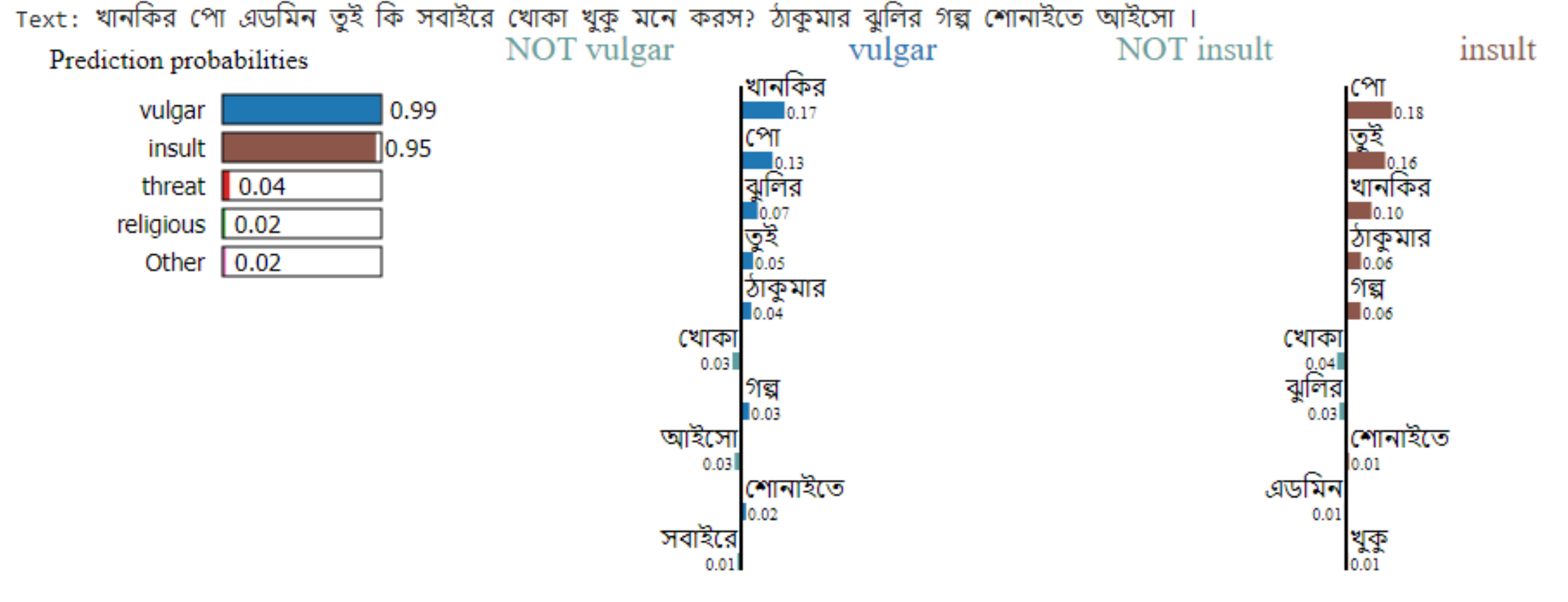}}
\caption{The explanation of the predictions made by the multi-label classifier along with top-2 class wise interpretation of word feature importance.}
\label{fig:toxic-lime-2}
\vspace{-5mm}
\end{figure*}
The performance of the CNN-BiLSTM with Attention mechanism model at all classification thresholds can be visualized in figure \ref{fig:roc-multi}. Furthermore, we can also observe in figure \ref{fig:toxic-lime-2} how the model categorizes a sample Bengali toxic comment into both \textit{vulgar} and \textit{insult} classes with $99\%$ and $95\%$ probability respectively considering ten important word features among which both classes are emphasized by four common word features. During experimentation, we observed that the dataset used in this study contains plenty of toxic words which are not present in the pretrained word embedding models like GloVe, fasttext, word2vec and the missing words are represented using null values which may produce sub-optimal feature representation for context understanding. Given the significance of the contextual meaning that a toxic comment can have, we have incorporated Bangla BERT to get the word embedding for feature representation. From table \ref{tab:dataset}, it is evident that the data distribution per toxic class is not equal which may produce bias towards particular classes. To justify this claim, we plot the confusion matrix in Figure \ref{fig:visualization_graph} for the four toxic categories that have lower f1-scores. 

\section{Conclusion}
In recent years, deep learning has achieved great success in various natural language processing tasks. In this paper, we have focused on the task of Bengali multi-labeled toxic comments classification. We have proposed a deep learning model based pipeline to effectively capture the dependencies among different labels and improve the classification accuracy that outperformed a fine-tuned pre-trained transformer model in both binary and multi-label classification tasks. Experimental results indicate that our proposed method is able to learn the complex structure of the Bengali language and capture important features for classification which can be visualized by using a text explainer framework. We leave extensive experimentation using transfer learning and fine tuning several pre-trained Bengali transformer models by increasing the data instances per class as future work.

\end{document}